\documentclass{article}
\usepackage{ijcai22}

\usepackage{times}
\usepackage{soul}
\usepackage{url}
\usepackage[hidelinks]{hyperref}
\usepackage[utf8]{inputenc}
\usepackage[small]{caption}
\usepackage{amsmath}
\usepackage{booktabs}

\usepackage{helvet}  
\usepackage{courier}  
\usepackage{graphicx} 
\usepackage{caption} 
\usepackage{subfigure}
\usepackage{amsfonts}
\usepackage{multirow}
\usepackage{color}

\usepackage{algorithm,algpseudocode}

\algnewcommand{\algorithmicforeach}{\textbf{for each}}
\algdef{SE}[FOR]{ForEach}{EndForEach}[1]
{\algorithmicforeach\ #1\ \algorithmicdo}
{\algorithmicend\ \algorithmicforeach}

\title{FATE-LLM: A Industrial Grade Federated Learning Framework for Large Language Models}

\author{
Tao Fan$^{1,2}$\footnote{Corresponding Author}\and
Yan Kang$^2$\and
Guoqiang Ma$^2$\and
Weijing Chen$^2$\and
Wenbin Wei$^2$\and
Lixin Fan$^2$\and
Qiang Yang$^{1,2}$\\
\affiliations
$^1$ Hong Kong University of Science and Technology, China \\
$^2$ WeBank, China 
\emails
tfanac@cse.ust.hk,
yangkang@webank.com,
zotrseeewma@webank.com,
weijingchen@webank.com,
sagewei@webank.com,
lixinfan@webank.com,
qyang@cse.ust.hk
}

\begin{document}

\maketitle

\begin{abstract}
Large Language Models (LLMs), such as ChatGPT, LLaMA, GLM, and PaLM, have exhibited remarkable performances across various tasks in recent years. However, LLMs face two main challenges in real-world applications. One challenge is that training LLMs consumes vast computing resources, preventing LLMs from being adopted by small and medium-sized enterprises with limited computing resources. Another is that training LLM requires a large amount of high-quality data, which are often scattered among enterprises. 

To address these challenges, we propose FATE-LLM, an industrial-grade federated learning framework for large language models. FATE-LLM (1) facilitates federated learning for large language models (coined FedLLM); (2) promotes efficient training of FedLLM using parameter-efficient fine-tuning methods; (3) protects the intellectual property of LLMs; (4) preserves data privacy during training and inference through privacy-preserving mechanisms. We release the code of FATE-LLM at \url{https://github.com/FederatedAI/FATE-LLM} to facilitate the research of FedLLM and enable a broad range of industrial applications.

\end{abstract}

\section{Introduction} 
In recent few years, the advent of large language models (LLMs) ~\cite{yang2023harnessing,zhou2023comprehensive} has been reshaping the field of artificial intelligence. In particular, the most advanced LLMs, such as ChatGPT~\cite{Chatgpt}, GPT-4~\cite{Gpt-4}, and PaLM~\cite{chowdhery2022palm} that boast billions of parameters have gained considerable attention due to their remarkable performance in a variety of natural language generation tasks. Many open-sourced LLMs with high performance have been released, and the public's enthusiasm for research and application of LLMs has been stimulated.
\begin{figure}[!ht]
  \centering
  \includegraphics[width=.35\textwidth]{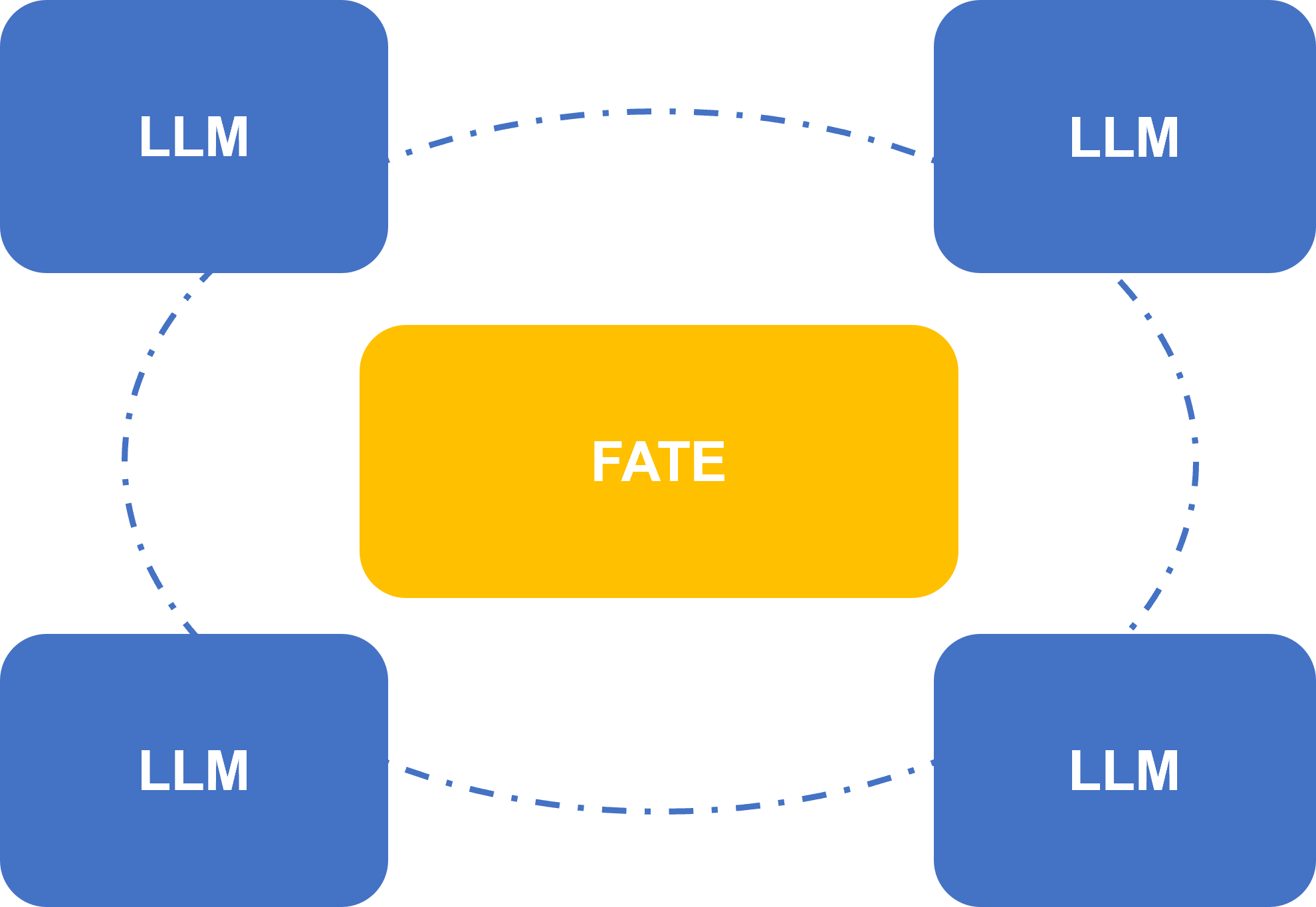}
  \caption{\textbf{Large Language Models are federated on FATE}. 
  }
  \label{FATE-LLM shows}
\end{figure}
 
However, grounding LLMs in real-world applications faces many challenges. The two main challenges are (i) training LLMs consumes vast computing resources, which prevents LLMs from being adopted by small and medium-sized companies with limited computing resources; (ii) training LLMs requires a large amount of public data, which may run out soon~\cite{villalobos2022will}.

Federated learning (FL)~\cite{mcmahan2017communication} \cite{yang2019federated}, a privacy-preserving collaborative machine learning paradigm, is a promising approach to deal with these two challenges. For one thing, FL enables many companies with different computing resources to collaboratively train powerful machine learning models such that the computational burden of training large models can be alleviated. For another, massive high-quality data are scattered among companies that are typically isolated from each other, and FL can exploit these data silos in a privacy-preserving way. 

In this work, we propose FATE-LLM, built upon FATE (Federated AI Technology Enabler)~\cite{liu2021fate}, to facilitate federated learning for large language models. More specifically, FATE-LLM (1) enables federated learning for both homogeneous and heterogeneous large language models (FedLLM); (2) promotes efficient training of FedLLM through parameter-efficient fine-tuning methods, such as LoRA~\cite{hu2021lora} and P-Tuning-v2~\cite{liu2021p}; (3) protects the intellectual property of LLMs using federated intellectual property protection approach~\cite{li2022fedipr}; (4) protects data privacy during training and inference through privacy-preserving mechanisms. We release the code of FATE-LLM at \url{https://github.com/FederatedAI/FATE-LLM} to promote the research of FedLLM and enable a broad range of industrial applications. 


\section{Related Work}
In this section, we briefly review related work regarding large language models and federated learning. 

\subsection{Large Language Models}
The advancements in large language models(LLMs) have led to significant advances in a variety of NLP tasks. A great example of LLMs application is ChatGPT\cite{Chatgpt}. ChatGPT is fine-tuned from the generative pretrained transformer GPT-3.5, which was trained on a blend of text and code. ChatGPT applies reinforcement learning from human feedback (RLHF), which has become a promising way to align LLMs with a human’s intent. LLMs are generally divided into two categories: encoder-decoder or encoder-only large language models and decoder-only large language models \cite{yang2023harnessing}. Bert \cite{devlin2018bert} is the representative of encoder-only large language models. GPTs \cite{radford2018improving} is the representative of decoder-only large language models. At the early stage of LLMs development, decoder-only LLMs were not as popular as encoder-only and encoder-decoder LLMs. However, after 2021, with the introduction of GPT-3 \cite{brown2020language}, decoder-only LLMs experienced a significant boom. At the same time, after the initial explosion brought about by BERT \cite{devlin2018bert}, encoder-only LLMs gradually began to fade away. Recently, many decoder-only LLMs have been released, such as LLaMA \cite{touvron2023llama}, OPT \cite{zhang2022opt}, PaLM \cite{chowdhery2022palm}, and BLOOM \cite{scao2022bloom}. These LLMs demonstrated reasonable few-/zero-shot performance via prompting and in-context learning.

\subsection{Federated Learning}
Federated learning (FL) \cite{mcmahan2017communication} \cite{yang2019federated,liu2022vertical} is a distributed machine learning paradigm that enables clients (devices or organizations) to train a machine learning model collaboratively without exposing clients' data.  Unlike traditional centralized machine learning techniques, data are fixed locally rather than being gathered in a central server, which exists many of the systemic privacy risks and costs \cite{kairouz2021advances}. Hence, FL is a promising approach to deal with this data isolation challenge.  To enhance data privacy, federated learning uses a variety of secure computing protocols. The most popular protocols are Homomorphic Encryption (HE) \cite{paillier1999public}, Multi-Party Computation(MPC) \cite{shamir1979share} \cite{damgaard2012multiparty}, and Differential Privacy (DP) \cite{dwork2014algorithmic}. In recent years, the literature has presented various algorithms in the FL setting. \cite{hardy2017private} proposed vertical logistic regression (VLR) using homomorphic encryption (HE) to protect data privacy. \cite{chen2021homomorphic} further enhanced the privacy-preserving capability of VLR by employing a hybrid strategy combining HE and secret sharing (SS). \cite{cheng2021secureboost} proposed the SecureBoost, a VFL version of XGBoost, that leverages HE to protect the parameters exchanged among parties. \cite{kang2022privacy} applied a semi-supervised learning method to estimate missing features and labels for further training. \cite{mcmahan2017communication} proposed Secure Aggregation to enhance data protection.

\section{FATE-LLM System Design}

We introduce the FATE-LLM system, including its components, architecture, and roadmap. 

\subsection{Overview of FATE-LLM system}

\begin{figure}[!ht]
  \centering
  \includegraphics[width=.49\textwidth]{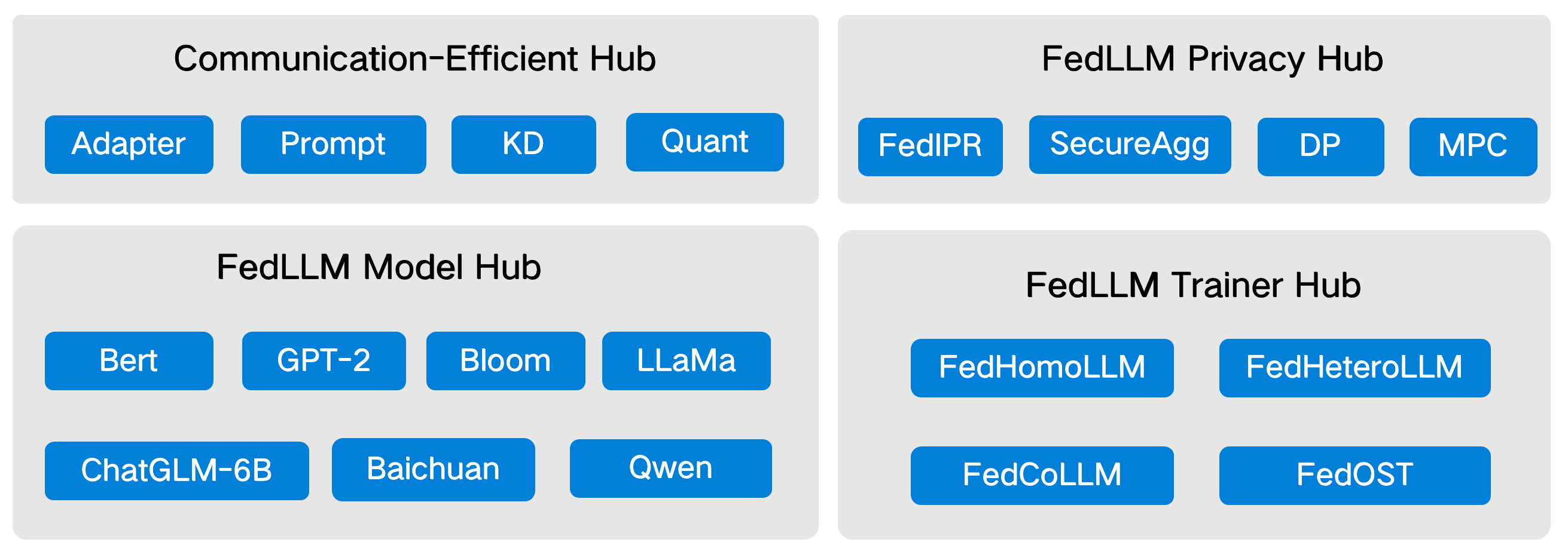}
  \caption{\textbf{Components of the FATE-LLM system}. 
  }
  \label{FATE-LLM overview}
\end{figure}

FATE-LLM\footnote{FATE-LLM was open-sourced in April 2023 in the FATE Community and is running on the infrastructure of FATE.} was open-sourced as a submodule of FATE, and it contains three components: Communication-Efficient Hub, FedLLM Model Hub, and FedLLM Privacy Hub. Figure \ref{FATE-LLM overview} overviews the FATE-LLM system.

\textbf{The Communication-Efficient Hub} integrates a variety of communication-efficient methods into FedLLM to reduce the communication cost for training LLMs, including parameter-efficiency fine-tuning (PEFT)~\cite{zhang2022federated} methods (e.g., Adapter Tuning \cite{cai2022autofednlp} and Prompt Tuning \cite{zhao2022reduce}, Knowledge Distillation(KD)~\cite{wu2022communication}, and Model Quantization~\cite{zhang2018lq}. More specifically, \cite{zhang2022federated} proposed PETuning methods that can reduce the communication overhead by \textbf{$1\sim 2$} orders of magnitude under the FL setting compared with full fine-tuning. They also found that PETuning methods can bring down local model adaptation costs for clients in FL systems. These results imply that FL clients (e.g., devices) with limited storage capacity can benefit from PETuning methods since these methods enable sharing an LLM across different tasks and maintaining a few parameters for each task, reducing the storage requirement. 

\begin{figure*}[ht]
\vspace{-10pt}
    \centering
    \subfigure[\textbf{FedHomoLLM} (Federated homogeneous LLMs): Clients have LLMs with the same architecture leverage PEFT to train their LLMs.]{
        \includegraphics[width=0.47\textwidth]{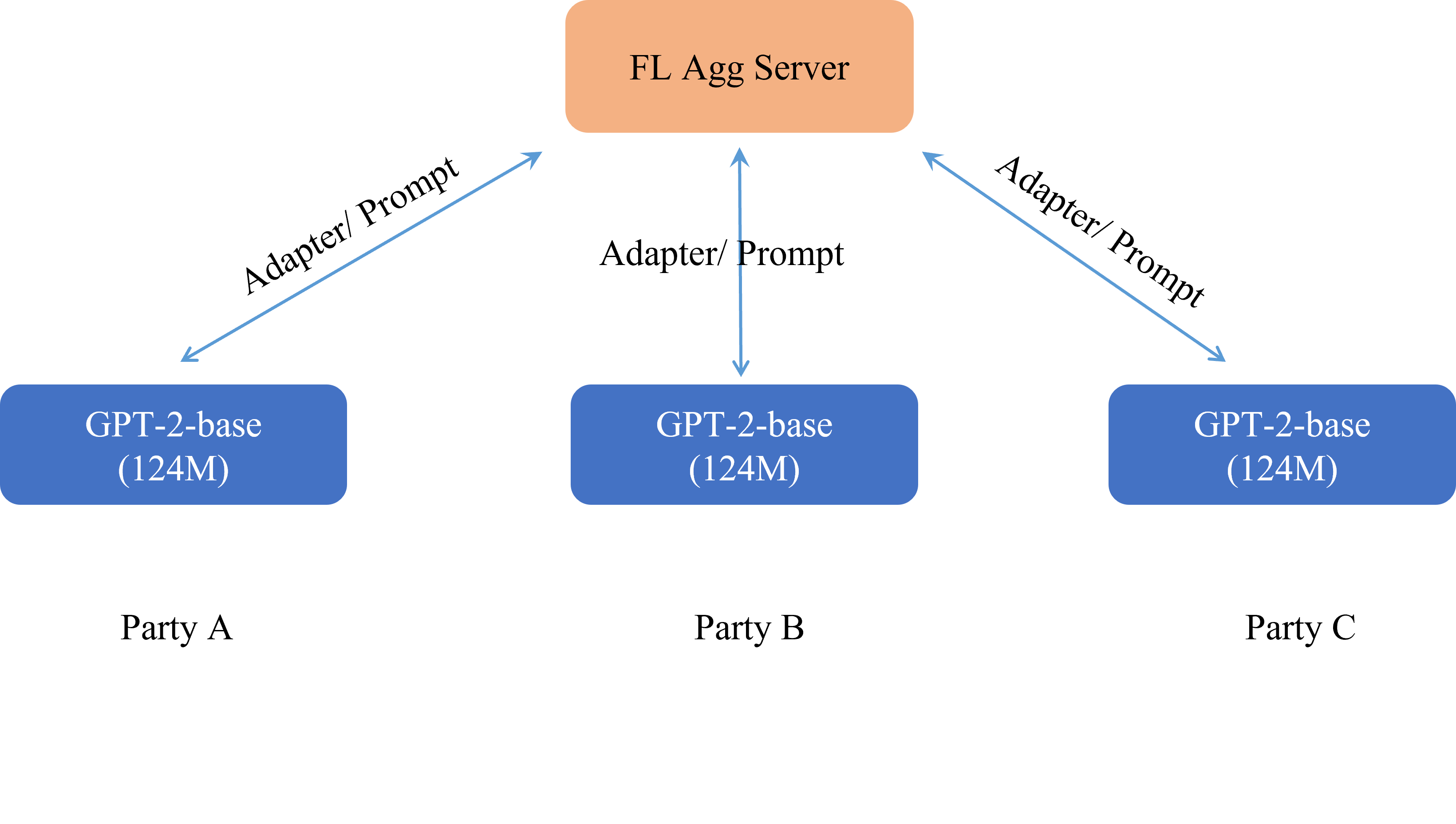} 
        \label{fig:fedhomollm}
    }
     \hfill
    \subfigure[\textbf{FedHeteroLLM} (Federated Heterogeneous LLMs): Clients have LLMs with different architecture leverage knowledge distillation and PEFT to train their LLMs.]{
        \includegraphics[width=0.47\textwidth]{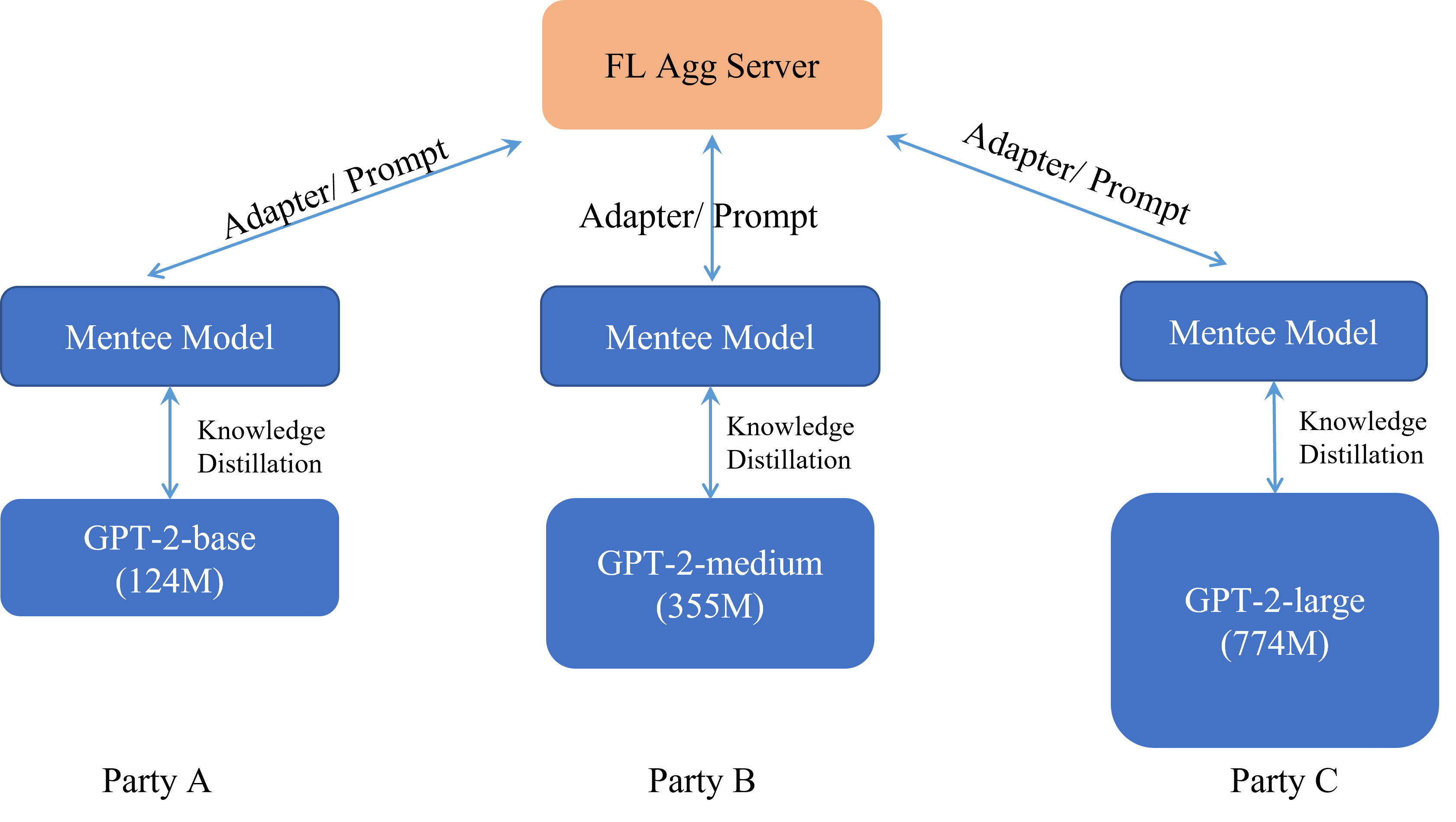} 
        \label{fig:fedheterollm}
    }
    \subfigure[\textbf{FedCoLLM} (Federated Co-tuning LLMs): Not only clients but also the server owns LLMs. They leverage PEFT and knowledge distillation to fine-tune their LLMs.]{
        \includegraphics[width=0.48\textwidth]{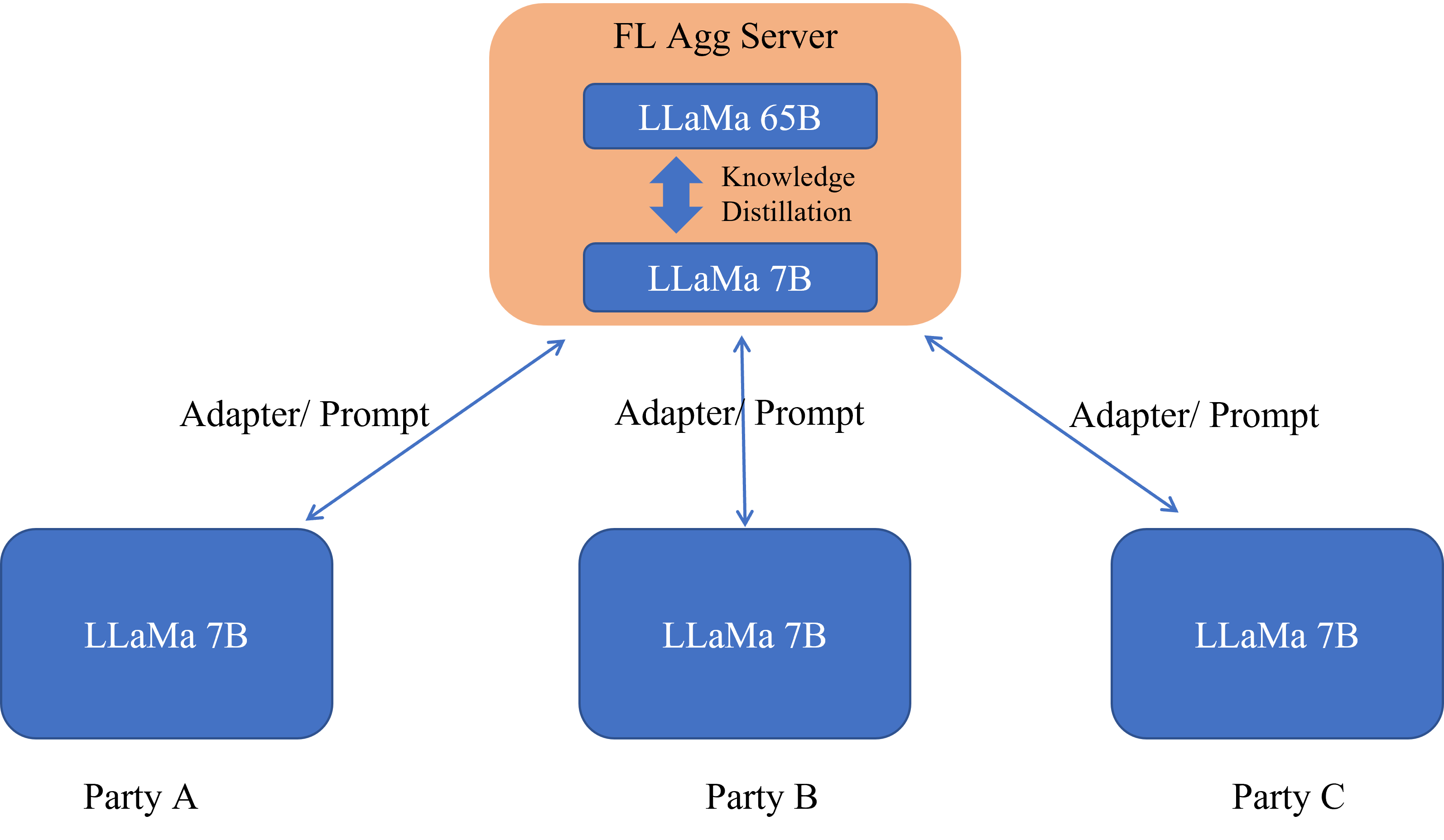} 
        \label{fig:fedcollm}
    }
     \hfill
    \subfigure[\textbf{FedOST} (Federated OffSite-Tuning): Clients transfer their knowledge to the LLM hosted by the server through offsite-tuning in a federated way.]{
        \includegraphics[width=0.47\textwidth]{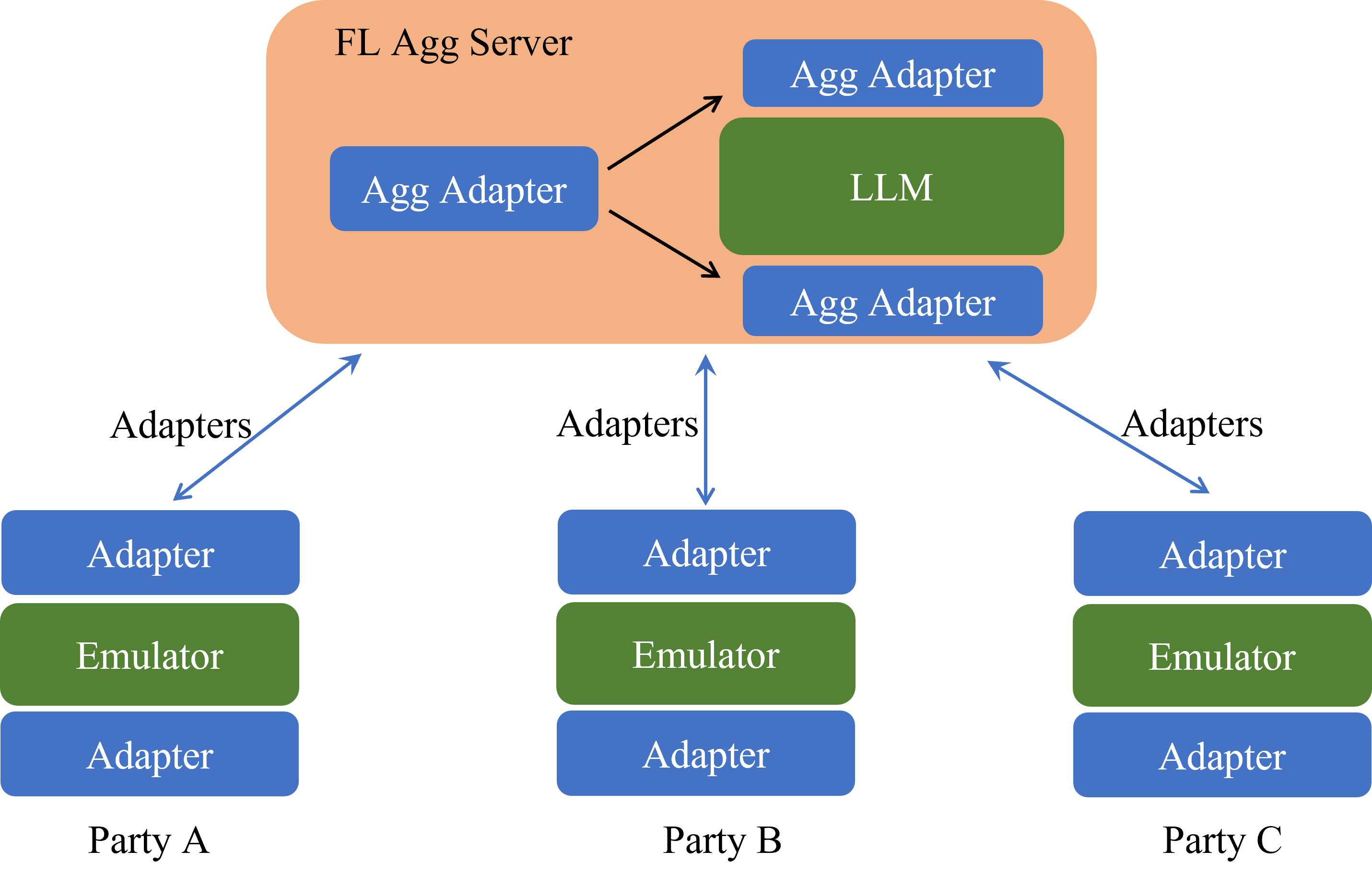} 
        \label{fig:fedost}
    }
    \caption{FATE-LLM Trainers. FATE-LLM offers four trainers for four different federated LLM learning scenarios.}
    \label{fig:fate-llm}
    \vspace{-10pt}
\end{figure*}

\textbf{The FedLLM Model Hub} integrates a variety of mainstream LLMs, including BERT~\cite{devlin2018bert}, GPTs~\cite{radford2018improving}, ChatGLM-6B~\cite{du2022glm}, LLaMA~\cite{touvron2023llama}, BLOOM~\cite{scao2022bloom}, and Baichuan~\cite{yang2023baichuan}. These LLMs have different architectures and sizes and can be applied in different scenarios.

\textbf{The FedLLM Trainer Hub} offers a variety of training methods for different federated LLMs learning scenarios, including FedHomoLLM, FedHeteroLLM, FedCoLLM, and FedOST.

In FL, clients may have sufficient computing resources to train LLMs of the same size. However, in many heterogeneous scenarios, clients are likely to have quite different computing or data resources so that they can afford to train LLMs of quite different sizes. FATE-LLM offers Federated Homogeneous LLMs (FedHomoLLM) and Federated Heterogeneous LLMs (FedHeteroLLM) to support both scenarios. FedHomoLLM leverages PEFT techniques to train clients' LLMs with the same architecture and size (illustrated in Figure \ref{fig:fedhomollm}). FedHeteroLLM leverages knowledge distillation (KD)~\cite{shen2020federated} and PEFT techniques to deal with the FL scenario where FL clients own LLMs of different sizes (illustrated in Figure \ref{fig:fedheterollm}). Specifically, each client in FedHeteroLLM leverages KD to learn a mentee model from its local pre-trained LLM. Then, all clients send adaptor or prompt parameters to the server for secure aggregation. Next, the server dispatches the aggregated model to all clients for the next round of training. 

Initializing clients with an LLM distilled from a larger one hosted by the server enables federated LLMs to obtain a better global model more efficiently than starting clients' models from random initialization~\cite{wang2023can}. On the other hand, the domain knowledge captured by clients' local LLMs allows the server's larger LLM to continue to evolve. FATE offers the FedCoLLM (Federated Co-tuning LLM) framework to co-evolve the LLMs of the server and clients. Figure \ref{fig:fedcollm} illustrates the FedCoLLM. Specifically, in FedCoLLM, each client having a LLaMa-7B model conducts federated learning applying PEFT techniques. On the server side, the server distills the knowledge between its LLaMa-65B model and the aggregated LLaMa-7B mode to co-evolve models on both sides.


\cite{xiao2023offsite} proposed Offsite-Tuning, a privacy-preserving and efficient transfer learning framework that can adapt an LLM to downstream tasks without access to the LLM's full weights. More specifically, in Offsite-Tuning, the server sends two adaptors and an emulator of its LLM to a client, which in turn finetunes adaptors with the help of the frozen emulator using its domain-specific data. Next, the client sends adaptors back to the server, which then plugs them into its LLM to form an adapted LLM for the client. The Offsite-Tuning has the potential to protect the client's data privacy and the server's model property.

FATE-LLM offers the FedOST~(Federated OffSite-Tuning) that extends the Offsite-Tuning framework to the federated learning setting (see Figure \ref{fig:fedost}). In FedOST, multiple clients collaboratively train two global adaptors that adapt the LLM to all clients. FedOST brings two additional benefits than Offsite-Tuning: (1) FedOST enhances data privacy by adopting secure aggregation, and (2) it adapts an LLM to clients that did not even participate in the FL because of the generalization of the FL global model.

\begin{figure}[!h]
  \centering
  \includegraphics[width=.48\textwidth]{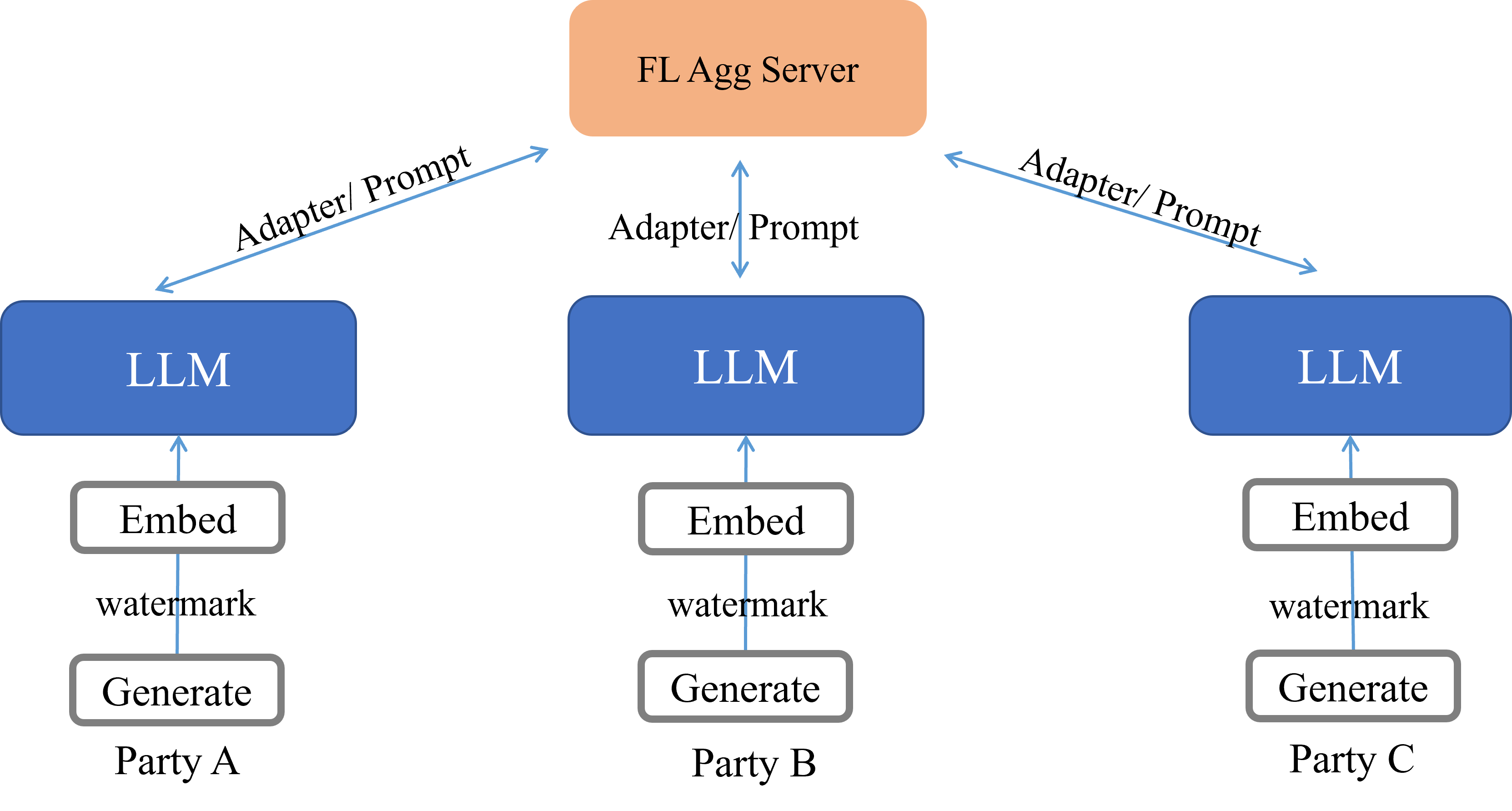}
  \caption{\textbf{FedIPR!\protect\cite{li2022fedipr}}. Private watermarks are generated and embedded into the trainable parameters (i.e., adaptors or prompts) of local large language models. Then, trainable parameters are aggregated through FedAvg.}
  \label{fig:fedipr}
\end{figure}

\textbf{The FedLLM Privacy Hub} integrates various privacy and security protection technologies, including federated intellectual property protection (FedIPR)~\cite{li2022fedipr}, secure aggregation (SecureAgg) ~\cite{mcmahan2017communication}, Differential Privacy (DP) and Multi-Party Computation (MPC) to protect data privacy and model security. Specifically, FedIPR~\cite{li2022fedipr} proposed a federated deep neural network ownership verification scheme that enables private watermarks to be embedded into private DNN models during FL training (see Figure \ref{fig:fedipr}) such that each client can independently verify the existence of embedded watermarks and claim its ownership of the federated model without disclosing private training data and watermark information. FedIPR can be applied to FedLLM to verify the IP ownership of the federated LLMs. SecureAgg, DP, and MPC can be applied to FedLLM during training and fine-tuning to protect clients' data privacy.  

\begin{table*}[!ht]
\centering
\begin{tabular}{c|c|c|c|c}
\hline
\multicolumn{1}{c}{Metrics} & \multicolumn{1}{c}{LoRA
Federated} & \multicolumn{1}{c}{LoRA
Centralized
} & \multicolumn{1}{c}{LoRA Client-1} & \multicolumn{1}{c}{LoRA Client-2} \\
\hline
\hline

Rouge-1 & 32.331  & 32.384     & 31.824    & 31.764     \\ \hline
Rouge-2 & 7.740  & 8.150  & 7.849   & 7.765\\ \hline
Rouge-$l$ & 25.600 & 25.830  & 25.408 & 25.404    \\ \hline
BLEU-4   & 8.344 & 8.730 & 8.340 & 8.366     \\ \hline

\end{tabular}
\caption{FedLLM fune-tuning ChatGLM-6B using LoRA.}
\label{lora exp}
\end{table*}

\begin{table*}[!ht]
\centering
\begin{tabular}{c|c|c|c|c}
\hline
\multicolumn{1}{c}{Metrics} & \multicolumn{1}{c}{P-Tuning-v2
Federated} & \multicolumn{1}{c}{P-Tuning-v2
Centralized
} & \multicolumn{1}{c}{P-Tuning-v2 Client-1} & \multicolumn{1}{c}{P-Tuning-v2 Client-2} \\
\hline
\hline

Rouge-1 & 32.227 &	32.184	& 31.362	& 31.18     \\ \hline
Rouge-2 & 7.644	& 8.048	& 7.472	& 7.478 \\ \hline
Rouge-$l$ & 25.853 &	26.010	& 25.454	& 25.227    \\ \hline
BLEU-4   & 8.490 &	8.851	& 8.329	& 8.221     \\ \hline
\end{tabular}
\caption{FedLLM fine-tuning ChatGLM-6B using P-Tuning-v2.}
\label{ptuningv2 exp}
\end{table*}

\subsection{Architecture of FATE-LLM}
\begin{figure}[!hb]
  \centering
  \includegraphics[width=.48\textwidth]{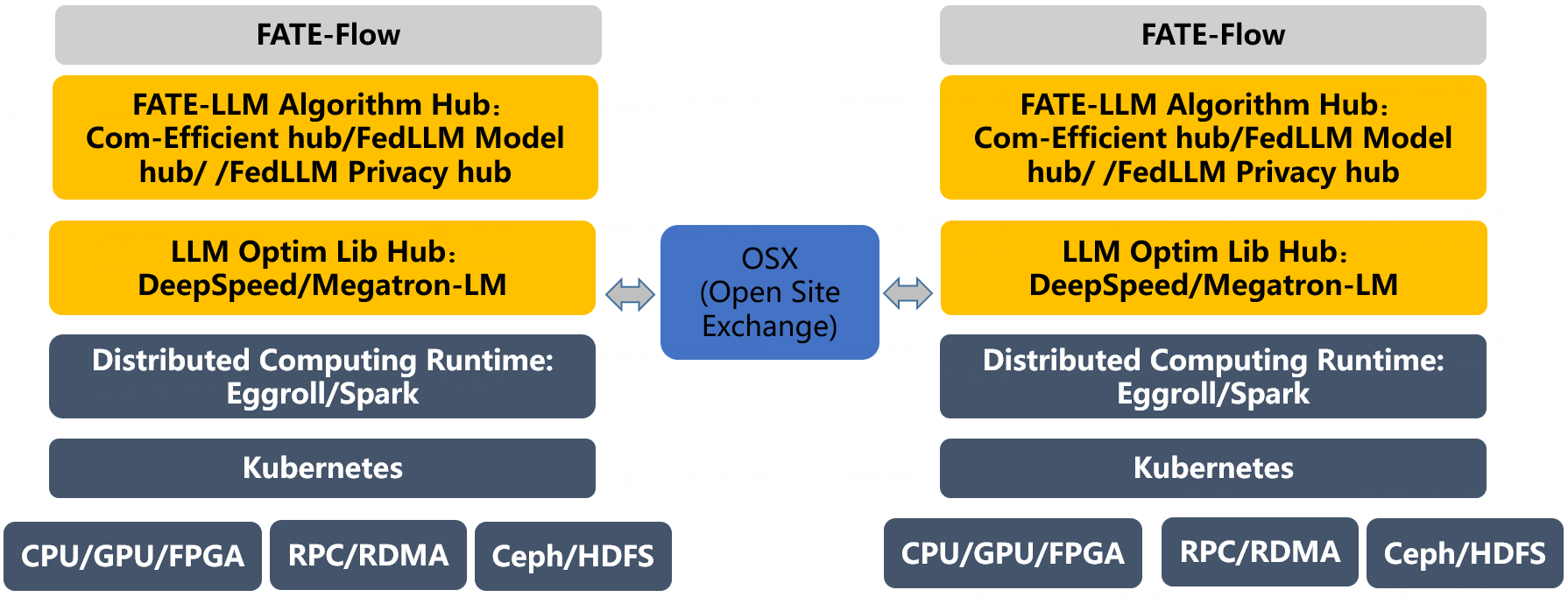}
  \caption{\textbf{Architecture of the FATE-LLM system}. 
  }
  \label{FATE-LLM Arch}
\end{figure}

FATE-LLM is running on the infrastructure of FATE, which consists of FATE-Flow, Eggroll, and OSX as the main components. FATE-Flow is a  task scheduling engine for the multi-party federated learning end-to-end pipeline, Eggroll is the distributed computing engine, and OSX (open site exchange) is the multi-party federated communication engine. FATE-LLM Algorithm Hub and LLM Optim Lib Hub are tailored to perform FedLLM. FATE-LLM Algorithm Hub includes Communication-Efficient Hub, FedLLM Model Hub, and FedLLM Privacy Hub (see Figure \ref{FATE-LLM overview}). LLM Optim Lib Hub includes DeepSpeed and Megatron-LM. As of June 2023, FATE has integrated DeepSpeed into Eggroll, which can manage the GPUs cluster well and dispatch DeepSpeed LLMs tasks. Figure \ref{FATE-LLM Arch} shows the architecture of FATE-LLM.

\subsection{RoadMap of FATE-LLM}

\begin{figure}[!h]
  \centering
  \includegraphics[width=.48\textwidth]{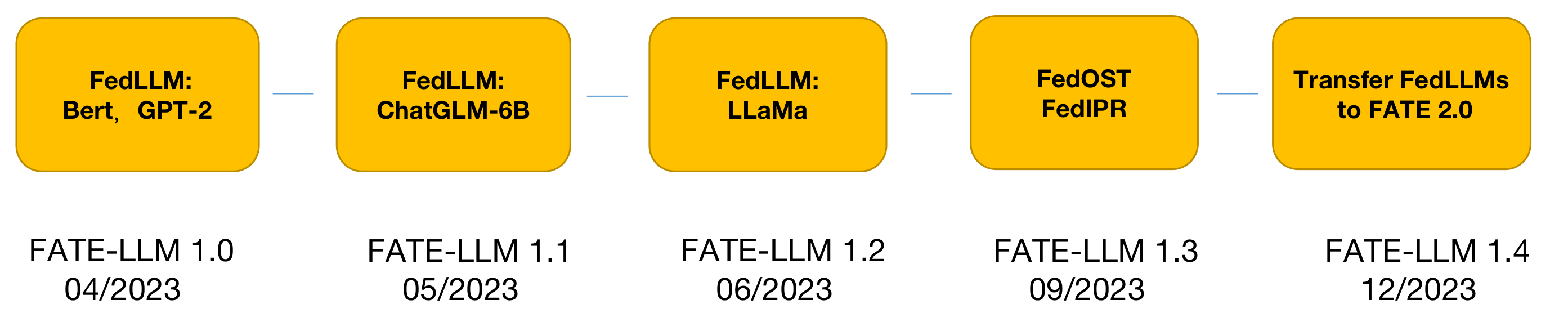}
  \caption{\textbf{RoadMap of FATE-LLM}. 
  }
  \label{FATE-LLM roadmap}
\end{figure}

We present the roadmap of FATE-LLM in Figure \ref{FATE-LLM roadmap}. As of June 2023, three versions of FTE-LLM have been released: FATE-LLM 1.0, FATE-LLM 1.1, and FATE-LLM 1.2. The three versions integrate Bert, GPT-2, ChatGLM-6B, and LLaMA, consecutively, and adopt FedIPR and privacy-preserving techniques to protect data privacy and model ownership. 

\section{Experiments}


We conduct experiments on the scenario in which each client owns a ChatGLM-6B~\cite{du2022glm} model, and all clients want to fine-tune their models collaboratively through federated learning. Since fine-tuning all parameters of ChatGLM-6B involves huge computational and communication costs, all clients leverage a PETuning method to only fine-tune a small portion of the ChatGLM-6B parameters through federated learning. 

\begin{figure}[!h]
  \centering
  \includegraphics[width=.48\textwidth]{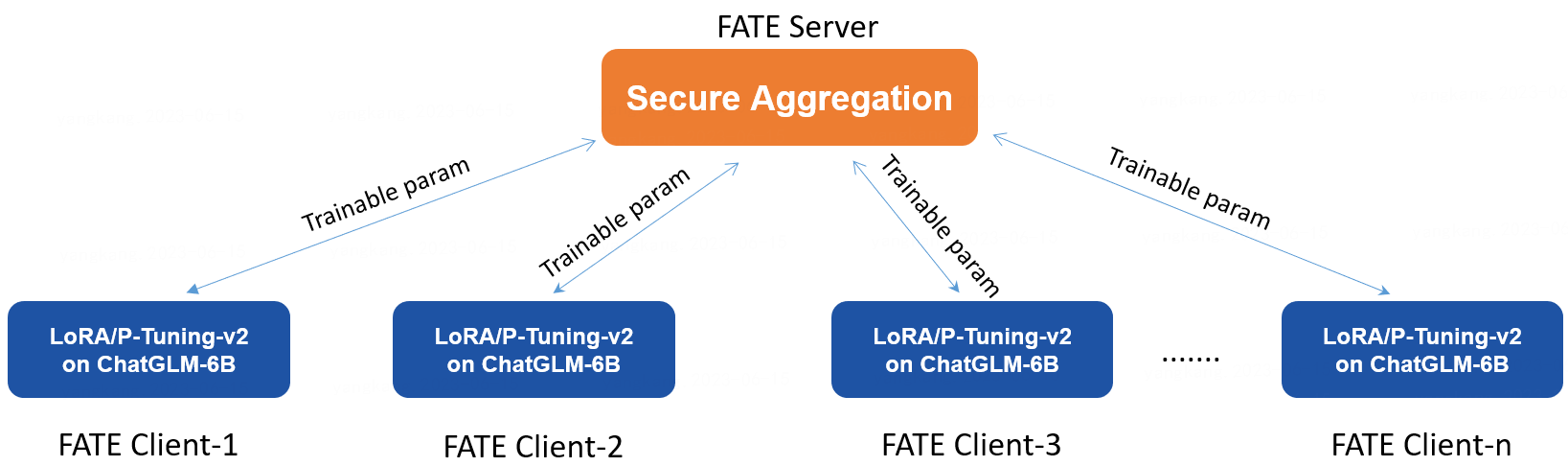}
  \caption{Multiple clients leverage LoRA or P-Tuning-v2 to fine-tine their local ChatGLM-6B models through federated learning. 
  }
  \label{FATE-LLM ChatGLM-6B}
\end{figure}

We leverage our FedLLM modules to conduct these experiments using both \textit{LoRA} \cite{hu2021lora} and \textit{P-Tuning-v2} \cite{liu2021p}. Figure \ref{FATE-LLM ChatGLM-6B} illustrates this scenario we conduct our experiments on.

\subsection{Experimental Setup}
We detail the experimental setup, including the dataset, FL setting, and baselines.

\textbf{Dataset and setting}. We conduct experiments on AdvertiseGen~\cite{shao2019long}, a dataset for advertising text generation. We simulate the FL setting with 2 clients and randomly split the AdvertiseGen dataset such that each client has 57K samples. Each client is assigned 8 NVIDIA V100 and trained on DeepSpeed. We set the FL training epoch to 5 and run the experiments in the LAN network environment. 

\textbf{Baselines}. We adopt two types of baselines. One is \textit{centralized}, in which data of all clients are centralized to conduct fine-tuning (either LoRA or P-Tuning-v2) on a ChatGLM-6B model. The another is that each client uses local data to fine-tune its local ChatGLM-6B model. 

\textbf{Evaluation metrics}. We adopt Rouge-1, Rouge-2, Rouge-$l$ \cite{lin2004rouge} and BLEU-4 \cite{papineni2002bleu} to evaluate the performance of fine-tined LLMs.


\subsection{Experiment Results} \label{sec:exp}

\subsubsection{Model Performance}
The experimental results for FedLLM using LoRA and P-Tuning-v2 are reported in Table \ref{lora exp} and Table \ref{ptuningv2 exp}, respectively, which show that LoRA Federated and P-Tuning-v2 Federated generally outperform their individual client counterparts across all performance metrics, demonstrating that federated learning help enhance the fine-tuning performance for each client. From Table \ref{lora exp} and Table \ref{ptuningv2 exp}, we also observe that the performance of LoRA and P-Tuning-v2 federated fine-tuning are generally worse than their centralized counterparts across all performance metrics, indicating that there has room to improve federated fine-tuning methods.


\subsubsection{Communication Cost}
We investigate the communication cost for FedLLM using LoRA and P-Tuning-v2 in terms of the size of parameters to be fine-tuned. Table \ref{tab:model_size} reports the results, and it shows that FedLLM using LoRA consumes 0.058\% communication cost of FedLLM fine-tuning all parameters, while FedLLM using P-Tuning-v2 accounts for 0.475\% communication cost of FedLLM fine-tuning all parameters.

\begin{table}[!h]
\centering
\begin{tabular}{c|c|c}
\hline
\multicolumn{1}{c}{Methods} & \multicolumn{1}{c}{
Model Size (MB)} & \multicolumn{1}{c}{Param Percent (\%)}  \\
\hline
\hline

LoRA & 3.6 &  0.058 \\ \hline
P-Tuning-v2 & 29.3	& 0.475\\ \hline
Fine-tune All & 6173 &	100	\\ \hline

\end{tabular}
\caption{Comparison of communication cost for FedLLM fine-tuning all parameters of ChatGLM-6B, fine-tuning ChatGLM-6B using LoRA and P-Tuning-v2. Model Size denotes the size of parameters to be fine-tuned. Param Percent denotes the ratio of parameters to be fine-tuned to all parameters.}
\label{tab:model_size}
\end{table}

\section{Conclusions and Future Work}

We proposed FATE-LLM, an industrial-grade federated learning framework for large language models(FedLLM). As an open-sourced software, FATE-LLM encourages collaboration among the research and industry communities and expects to receive increasing feedback on its use. 

In the future, we may consider research directions: (1) reconcile LLMs of different model architectures during FL fine-tuning; (2) fine-tune private LLMs of one party using private data of another party without compromising the data privacy and model ownership; (3) protect the privacy of user prompts efficiently in the inference stage; (4) apply FedLLM to vertical federated learning~\cite{liu2022vertical}.


\bibliographystyle{named}
\bibliography{fate-llm}

\end{document}